\documentclass[lettersize,journal]{IEEEtran}
\usepackage{amsmath,amsfonts}
\usepackage{algorithmic}
\usepackage{array}
\usepackage[caption=false,font=normalsize,labelfont=sf,textfont=sf]{subfig}
\usepackage{textcomp}
\usepackage{stfloats}
\usepackage{url}
\usepackage{verbatim}
\usepackage{graphicx}
\usepackage{multirow}
\usepackage{makecell}
\usepackage{tabularx}
\usepackage{threeparttable}
\usepackage{color}

\usepackage{amssymb}
\usepackage{pifont}
\newcommand{\cmark}{\ding{51}}%
\renewcommand{\paragraph}[1]{
 \textbf{#1}~}

\def\BibTeX{{\rm B\kern-.05em{\sc i\kern-.025em b}\kern-.08em
    T\kern-.1667em\lower.7ex\hbox{E}\kern-.125emX}}
\usepackage{balance}
\begin{document}
\title{SDM-Car: A Dataset for Small and Dim Moving Vehicles Detection in Satellite Videos}
\author{Zhen Zhang, Tao Peng, Liang Liao, Jing Xiao, and Mi Wang
\thanks{This work was supported by National Natural Science Foundation of China (62371351,62202349). (\emph{Corresponding author: Jing Xiao.})\par
Z. Zhang, P. Tao and J. Xiao are with the School of Computer Science, Wuhan University, Wuhan 430072, China (e-mail: zhangzhent@whu.edu.cn, 2022282110111@whu.edu.cn, jing@whu.edu.cn).\par
L. Liao is with the School of Computer Science and Engineering, Nanyang Technological University, 50 Nanyang Avenue, Singapore 639798 (e-mail: liang.liao@ntu.edu.sg).\par
M. Wang is with the State Key Laboratory of Information Engineering in Surveying, Mapping and Remote Sensing, Wuhan University, Wuhan 430072, China (e-mail: wangmi@whu.edu.cn).}}

\markboth{IEEE GEOSCIENCE AND REMOTE SENSING LETTERS}%
{}

\maketitle

\begin{abstract}
Vehicle detection and tracking in satellite video is essential in remote sensing (RS) applications. However, upon the statistical analysis of existing datasets, we find that the dim vehicles with low radiation intensity and limited contrast against the background are rarely annotated, which leads to the poor effect of existing approaches in detecting moving vehicles under low radiation conditions. In this paper, we address the challenge by building a \textbf{S}mall and \textbf{D}im \textbf{M}oving Cars (SDM-Car) dataset with a multitude of annotations for dim vehicles in satellite videos, which is collected by the Luojia 3-01 satellite and comprises 99 high-quality videos. Furthermore, we propose a method based on image enhancement and attention mechanisms to improve the detection accuracy of dim vehicles, serving as a benchmark for evaluating the dataset. Finally, we assess the performance of several representative methods on SDM-Car and present insightful findings. The dataset is openly available at https://github.com/TanedaM/SDM-Car.

\end{abstract}

\section{Introduction}
With the development of remote sensing technology, high-resolution satellite videos have emerged as one of the most essential carriers of continuous dynamic information for earth observation. Moving vehicle detection in satellite video technology, which aims at localizing and identifying vehicles in continuous motion from the satellite perspective, has grown widespread in several fields, such as digital city, smart traffic and intelligent surveillance. 

\begin{figure}
    \centering
    \includegraphics[width=0.95\linewidth]{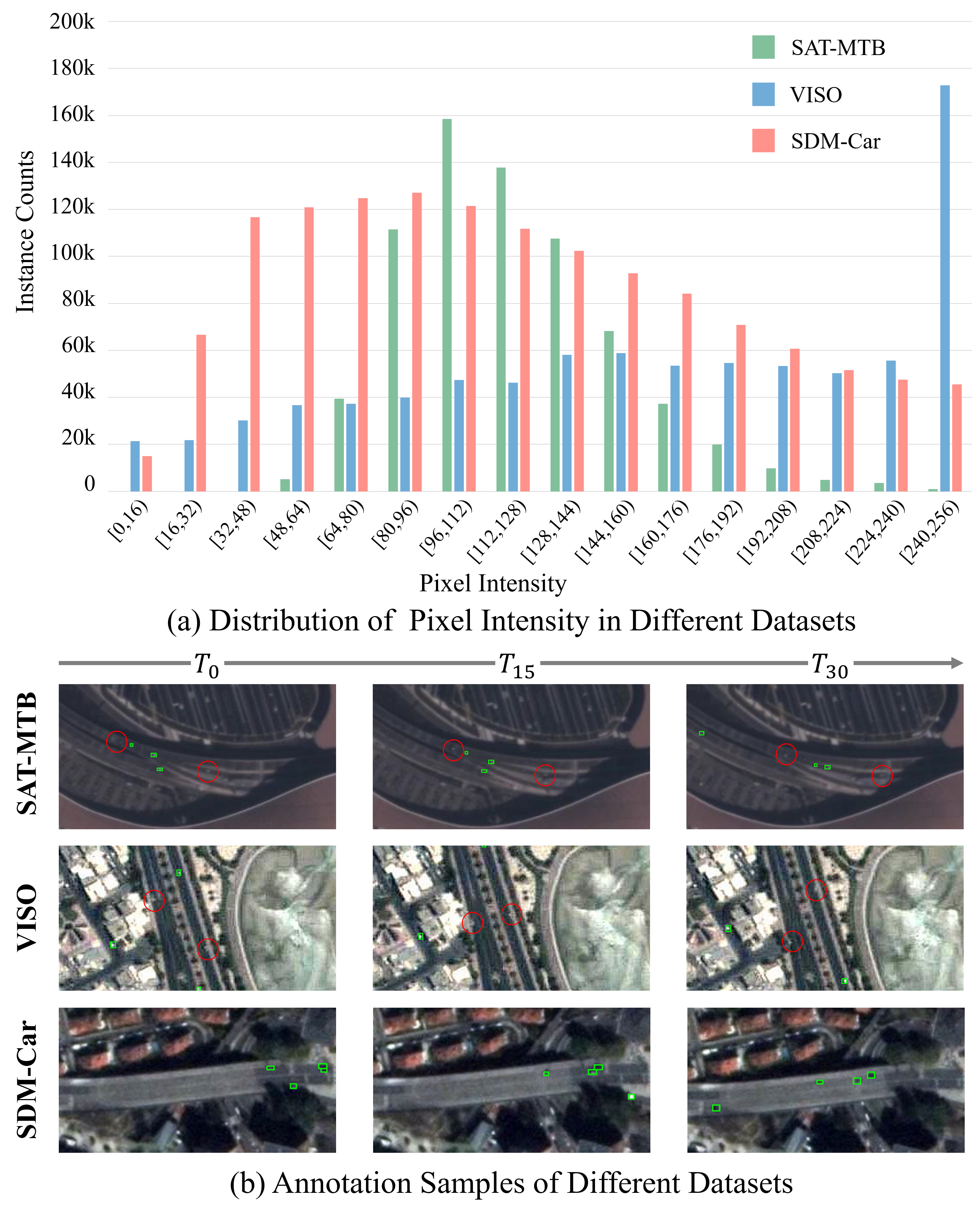}
    \vspace{-0.5cm}
    \caption{(a) Statistic analysis of the pixel intensity distribution of different datasets. The horizontal axis and vertical axis represent the pixel intensity of the center of the annotation box and the instance counts, respectively. (b) Visualization of annotations for different datasets. The green boxes are annotation boxes, while the red circles are unannotated vehicles. The interval between adjacent images is 15 frames. 
    These unannotated vehicles are difficult to distinguish due to the low pixel intensity, but their motion information is perceptible to the human eye.
    }
    \label{fig:intro}
    \vspace{-0.4cm}
\end{figure}

Object detection is a long-standing research field in computer vision. However, there are still challenges in remote sensing scenarios, such as extremely small moving targets, complex and diverse background noises, motion artifacts caused by imaging platforms, and so on. 
To solve these issues, traditional methods for moving vehicle detection in satellite videos can be divided into two categories, \textit{i.e.}, frame difference, and background subtraction. Frame difference detects moving objects by calculating per-pixel differences between consecutive frames~\cite{7881701}, whereas background subtraction estimates a static background and subtracts it from the current frame to obtain moving objects~\cite{Zhang_Jia_Hu_2020, 9037205}. However, these methods are easily influenced by pixel variations caused by non-static satellite platforms and changes in lighting and contrast.

In recent years, many researchers have dedicated to applying deep learning models for vehicle detection~\cite{LIAO202345}. Xiao \textit{et al.} \cite{9594855} proposed a two-stream detection network named DSFNet with the input of consecutive frames to detect moving objects in satellite videos. Feng \textit{et al.} \cite{10061452} proposed a semantic-embedded density adaptive network named SDANet. The integration of the road prior to the network enables the model to focus on regions of interest, thereby reducing false detection. 
More recently, Xu \textit{et al.} \cite{xu2024real} proposed a historical frame differential module based on YOLOv5 to improve the ability to distinguish blurred vehicles by utilizing the motion features of historical frames, and meet the real-time detection requirements. Liao \textit{et al.} \cite{liao2024advancing} proposed to utilize the correlation between the dynamic characteristics of vehicles and the background to better localize the vehicles.
However, these methods frequently fail to detect dim vehicles that lack sufficient contrast against the roads, which are probably black or dark brown cars.

On the one hand, the low pixel intensity of dim vehicles leads to tiny contrasts with the background, which makes them difficult to detect. On the other hand, we noticed that the imbalance in pixel intensity of annotations in existing datasets exacerbates this challenge. To demonstrate this problem, we conduct statistics on the latest datasets, SAT-MTB~\cite{li2023multi} and VISO~\cite{9625976}. Specifically, we take the pixel value at the center of the annotation box as the pixel intensity of the vehicle and count the number of vehicles in each pixel intensity interval. The results shown in Fig.~\ref{fig:intro} (a) demonstrate the lack of low pixel intensity samples in existing datasets. Going a step further, Fig.~\ref{fig:intro} (b) presents annotation sample sequences from these datasets. A careful observation reveals that these datasets tend to label obvious vehicles with high pixel intensity, while some imperceptible vehicles with low pixel intensity, marked with red circles in the figure, are frequently unannotated.

In this paper, we propose a dataset named \textbf{S}mall and \textbf{D}im \textbf{M}oving Cars (SDM-Car) with million-level annotations for dim vehicles, to advance the task of moving vehicle detection in satellite videos. 
The SDM-Car dataset, in comparison to the current datasets, provides a large number of precise labeling for moving vehicles of all pixel intensity, especially dim vehicles, which can offer a more accurate and complete benchmark for vehicle detection and tracking from satellite images. The instances in the SDM-Car dataset are characterized by various densities, small instances, and complex backgrounds, indicating that the dataset is more challenging and existing methods do not yield good results. In addition, we propose a benchmark method to address the problem of detecting dim vehicles, including an image enhancement module and attention mechanisms. In summary, the main contributions of this paper are as follows:

1) We construct the SDM-Car dataset for moving vehicle detection in satellite videos. To the best of our knowledge, SDM-Car is the first dataset with substantial annotations for dim vehicles, providing the community with a superior data resource.

2) We propose a new benchmark method with image enhancement and attention mechanisms to enhance the capacity to capture local weak motion, which improves the detection performance of dim vehicles.

3) We evaluate a set of representative algorithms of moving vehicle detection in satellite videos on our new dataset, which can serve as baselines for future research.

\section{SDM-Car DATASET}
\subsection{Data Collection}
The SDM-Car dataset comprises 99 satellite videos captured by the Luojia 3-01 satellite, which is an Internet-intelligent Remote Sensing Satellite developed by Wuhan University and China Spacesat Co., Ltd.~\cite{deren2022new}. The satellite videos are all obtained by the Luojia 3-01 video staring imaging mode, with a spatial resolution of 0.75m at 500km orbital altitude with a frame rate of 6/8 fps. All videos are in true color and last around 30 seconds.
These videos cover a variety of scenes, including deserts, urban areas, ports, rural landscapes, and forests, enhancing their adaptability to vehicle detection tasks in diverse and complex scenarios. 


\begin{table*}[h]
\caption{Overview of existing satellite video datasets for vehicle detection.}
\vspace{-2mm}
\centering
\label{tab:tab1}
\scalebox{1}{
\begin{tabular}{rcccccccc}
\hline
Dataset  & Year& Category & Videos &  Frames & Vehicle Instances & Fps & Spatial Resolution & Data Source \\ \hline
SatSOT~\cite{9672083} & 2022& 4 & 105 & 17,095& 27,664  &10/25 & 0.92-1m$^{*}$& Jilin-1, Skybox, Carbonite-2\\  
SAT-MTB~\cite{li2023multi} & 2023 & 2& 249 & 50,046& 704,552&10& 0.92m&Jilin-1\\ 
VISO~\cite{9625976} & 2022 & 4 & 47& 17,730& 837,688&10& 0.92m& Jilin-1\\  
\hline  
\textbf{SDM-Car} & 2024 & 1& 99 & 16,830& \textbf{1,469,948}&6/8& \textbf{0.75m}&Luojia 3-01\\ \hline 
&&&&&
\multicolumn{4}{c}{\scriptsize{Spatial resolution marked with $^*$ is extrapolated from the data source.}}\\
\end{tabular}}
\vspace{-5mm}
\end{table*}

\begin{figure}[tb]
    \centering
    \includegraphics[width=1\linewidth]{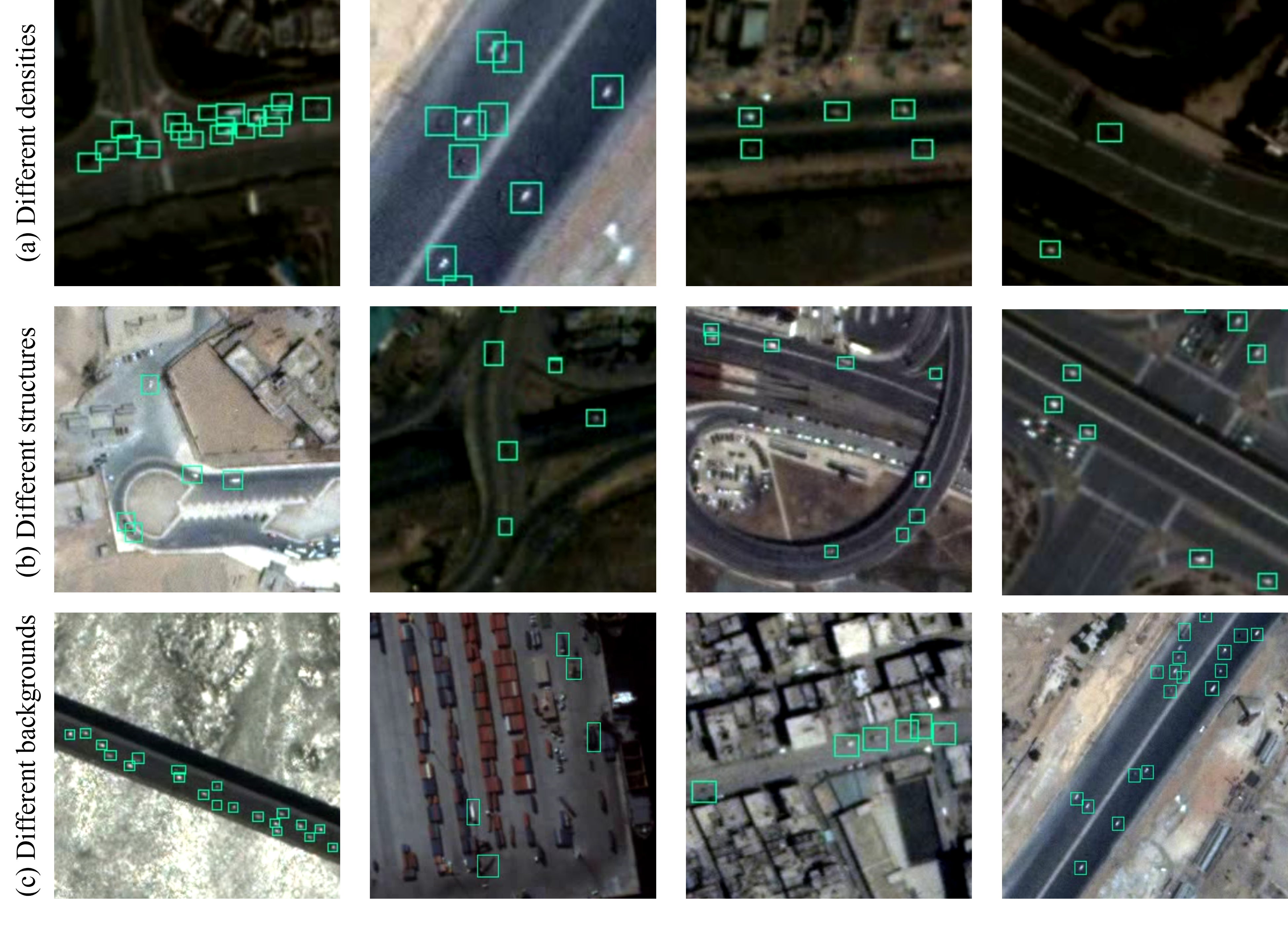}
    \vspace{-8mm}
    \caption{Example images and their annotations in the SDM-Car dataset. Our dataset shows diversity in terms of instance density, road structure, and background, and contains a large number of dim instances.}
    \label{fig.2}
    \vspace{-6mm}
\end{figure}

\subsection{Data Annotation}
The proposed SDM-Car dataset is dedicated to the detection of small and dim targets, with a sole category of vehicles. As vehicles are typically rigid and relatively small (normally less than 20 pixels), we annotate objects only using unoriented bounding boxes. The annotation includes the frame sequence number, object number, coordinates of the upper-left corner, width, and height of the bounding box. Additionally, since static dim vehicles are difficult to distinguish from the background even manually, our dataset exclusively focuses on moving vehicles, which means that static objects in the videos are not annotated. The original videos are segmented into multiple patches without overlap with a pixel resolution of 1920$\times$1080 for annotation.
We manually annotate the first frame of the video, while the subsequent frames are predicted by an annotation tool DarkLabel\footnote{https://github.com/darkpgmr/DarkLabel} and manually adjusted. Some examples of annotations are shown in Fig. \ref{fig.2}. In summary, we annotated 16,830 frames of 99 videos and obtained millions of annotations. 

The comparison of dataset sizes and other basic information is shown in Table \ref{tab:tab1}. 
We compare with existing satellite video datasets for vehicle detection, including SatSOT~\cite{9672083}, SAT-MTB~\cite{li2023multi} and VISO~\cite{9625976}. SatSOT is a single object tracking dataset with 105 sequences. SAT-MTB is a multi-task satellite video dataset with lots of annotated videos, which can be used for vehicle detection. VISO is a dataset specifically for annotating small targets such as vehicles.
Among these datasets, our dataset provides the best spatial resolution and largest number of annotated vehicle instances.

\subsection{Dataset Challenges}
The proposed SDM-Car dataset focuses on the detection of dim vehicles in satellite videos, and it is very challenging in terms of instance brightness, instance density, instance size, and complicated background. 


\subsubsection{Dim Instance} A significant advantage of the proposed dataset is that it includes a substantial number of dim instances. To demonstrate this, we counted the number of instances at different pixel intensities. Specifically, we pick the pixel value at the center of the connotation box as the pixel intensity of the instance and count the number of instances that fall within the non-overlapping pixel intensity intervals with a step size of 16. The results shown in Fig.~\ref{fig:intro} indicate that the proposed dataset contains a lot of dim instances and the distribution of pixel intensities is relatively uniform. Additionally, we also compared with other released satellite video datasets for vehicle detection. The mean pixel intensity of the SDM-Car dataset is 125.78, which is lower than 167.68 of VISO and 128.60 of SAT-MTB. Furthermore, although the mean pixel intensity of SAT-MTB is close to SDM-Car, the standard deviation of the proposed SDM-Car is 62.42, which is much higher than 31.50 of SAT-MAT. Overall, the proposed SDM-Car contains various instances of uniformly distributed brightness, including a large number of dim instances, which makes it hard to distinguish them from the background hampered by their low pixel intensity.


\subsubsection{Various Instance Density} 
\begin{figure}[tb]
    \centering
    \includegraphics[width=0.98\linewidth]{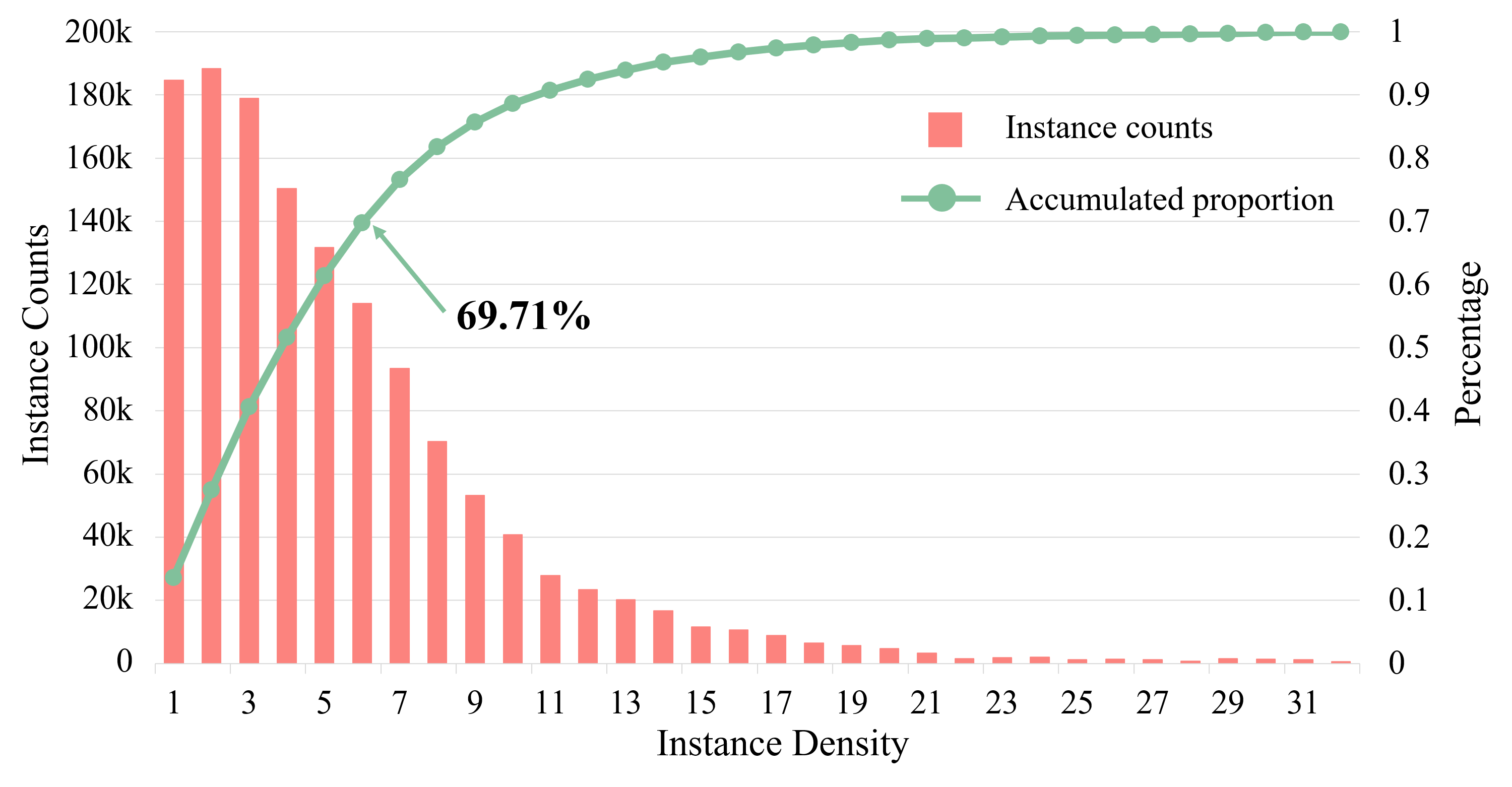}
    \vspace{-0.4cm}
    \caption{Distribution of instance density of in our dataset. We counted the number of instances in a 128$\times$128 patch as the approximation of the density. The histogram on the primary axis shows the number distribution of vehicles with different densities, and the curve chart on the secondary axis shows their accumulated proportions.}
    \label{fig:fig.3}
\end{figure}

We counted the number of instances per unit area in the proposed SDM-Car. Particularly, for statistical convenience, we split each frame into non-repeating patches with the pixel size of 128$\times$128, and count the number of instances whose center of the annotation box falls into the patch as the density. 
Fig.~\ref{fig:fig.3} illustrates the distribution of instance density. The histogram on the primary axis shows the number distribution of vehicles with different densities, and the curve chart on the secondary axis shows their accumulated proportions. The number of densely clustered vehicles (dispersed in patches with a density greater than 6) accounts for 30.29\% of the total, which may confuse the boundaries of vehicles and make accurate detection difficult.


\subsubsection{Small Instance} 
Due to the long photography distance of the remote sensing camera, the number of pixels occupied by the vehicle in the remote sensing image is very small. Although the spatial resolution of Luojia 3-01 is 0.75m, which is higher than the data sources of other datasets and the vehicle target is relatively larger in the image, 96.60\% of the instances is still not greater than 12 pixels in terms of width size, making the vehicle more difficult to be located.



\subsubsection{Complex background} As shown in Fig.~\ref{fig.2}, our dataset includes roads with various curvatures and structures, which makes the vehicle motion more complicated. In addition, our dataset also contains scenes of different styles, such as cities and deserts, bridges and ports, which puts higher requirements on the robustness to backgrounds.

\section{PROPOSED METHOD}
\label{sec:format}
\subsection{Framework Overview}

The challenge of detecting dim vehicles is that their radiation intensity is weak and the difference from the background is not obvious, which makes it difficult to distinguish them from the background and noise in static images. Therefore, we adopt the DSFNet presented in \cite{9594855} as our base model, as it utilizes motion information to detect vehicles. 

\begin{figure}
    \centering
    \includegraphics[width=1\linewidth]{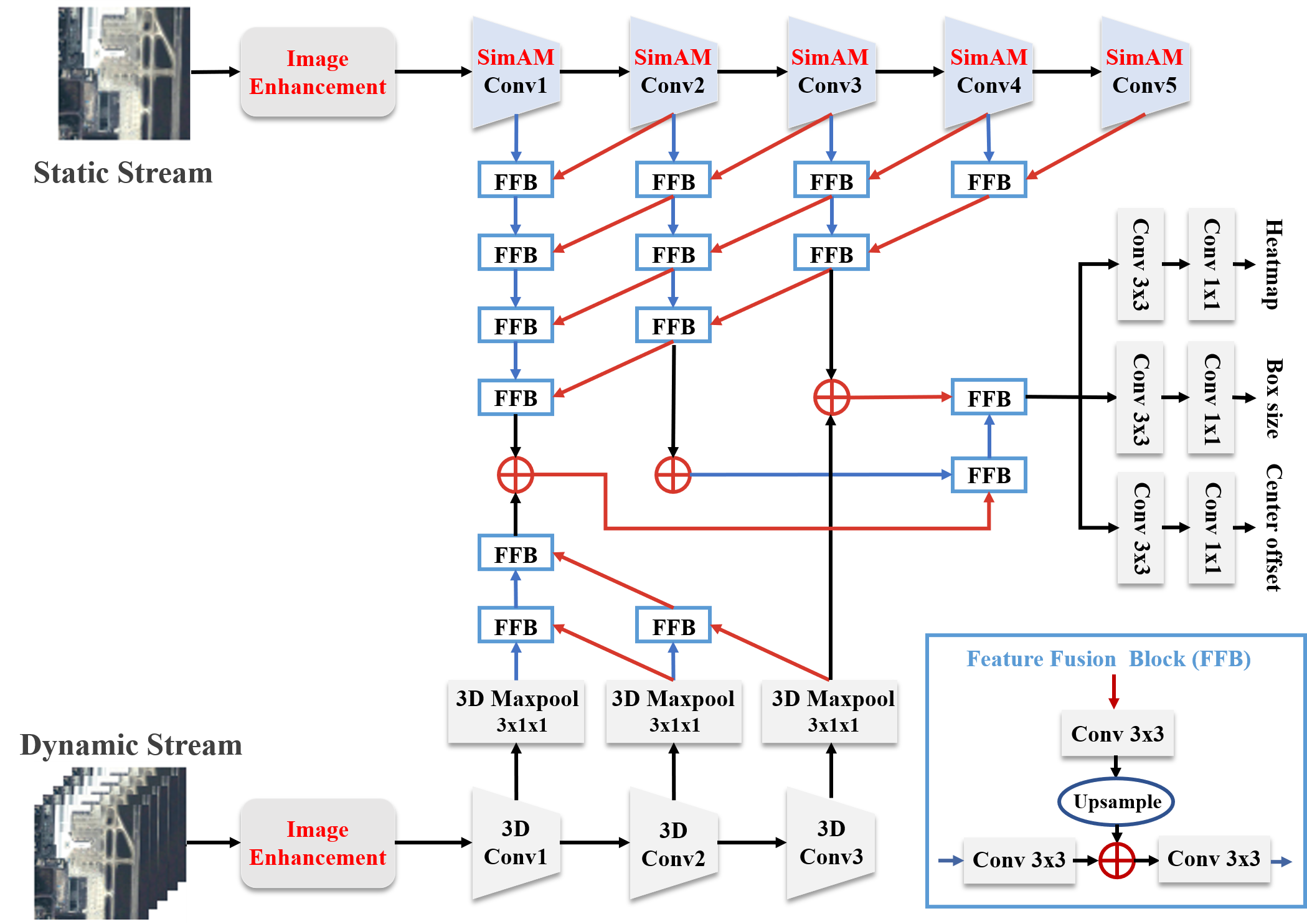}
    \caption{Overall architecture of the proposed method. The video frames are initially enhanced and then fed into a 2-D static stream (superior) and a 3-D dynamic stream (inferior). The SimAM module is incorporated into the 2-D static stream. 
    Features extracted by the aforementioned two streams are subsequently fused and input into the detection head, which yields the final detection outcomes.
    }
    \label{fig:method}
\end{figure}

The framework of the proposed method is shown in Fig. \ref{fig:method}. Specifically, the current frame and context frame sequence are fed into the static and dynamic streams, respectively, to generate feature representations. Those representations are subsequently fused and forwarded to the detection head, yielding the final detection results. 
In addition, we propose to utilize image enhancement and attention mechanisms to further improve the detection performance of dim vehicles. We augment the contrast and brightness of each frame before its input into the network to widen the intensity difference between the dim vehicle and the background at the pixel level. Besides, we incorporate the SimAM attention mechanism \cite{pmlr-v139-yang21o} into the feature extraction network to enhance the extraction of dim target features. 

\begin{figure}[h]
    \centering
    \includegraphics[width=0.85\linewidth]{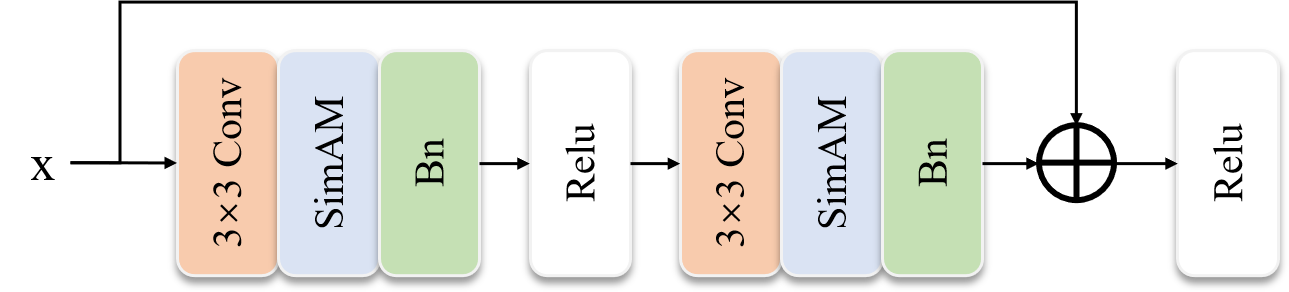}
    \caption{The structure of basicblock with SimAM. Bn: Batch Normalization. Relu: Rectified Linear Unit.}
    \label{fig:fig.2}
\end{figure}

\subsection{Image Brightness and Contrast Enhancement}
Because the proportion of pixels of dim vehicles in the image is minimal, the enhancement of the whole image has limited effects on the improvement of local contrast. Therefore, the contrast limited adaptive histogram equalization~\cite{reza2004realization} is adopted as our image enhancement module. 
Specifically, we divide the whole image into multiple patches with pixel size of 128$\times$128 evenly and perform histogram equalization independently in each patch. Finally, we calculate the bilinear interpolation between each adjacent patch to achieve a smooth transition between patches. By performing local contrast enhancement based on the image content in different areas, the contrast of dim vehicles is improved, making them easier to detect. Moreover, the overall quality of the image can also be maintained, avoiding problems such as over-enhancement and noise amplification.

\subsection{Attention Mechanism}
A significant amount of feature information in small objects will be lost after numerous downsampling and pooling layers, making it challenging to extract useful semantic feature information and detect small objects with accuracy.
To prompt the model to focus on dim vehicles, we integrate SimAM~\cite{pmlr-v139-yang21o}, an attention module, into the basic block of the feature extraction network. 
The upgraded module is illustrated in Fig. \ref{fig:fig.2}.
This module inhibits the interference of the background on the vehicle detection. In addition, by devising an energy function to compute attention weights, it dispenses with the necessity for an auxiliary sub-network to generate attention weights and steer clear of introducing supplementary parameters.

\section{EXPERIMENTAL RESULTS}
\label{sec:pagestyle}
\subsection{Experimental Setup}
We evaluate several representative baseline methods on our SDM-Car dataset, including three traditional methods, \textit{i.e.}, VIBE~\cite{barnich2009vibe}, GMM~\cite{friedman2013image}, KNN~\cite{elgammal2000non}, and two learning-based methods, \textit{i.e.}, HB-YOLO~\cite{yu2023hb}, DSFNet~\cite{9594855}.
To quantify the performance of these methods on our SDM-Car dataset, we adopt three evaluation metrics, including \textit{Recall}, \textit{Precision}, and \textit{$F_1$ Score}. The vehicle is considered to be properly detected if the predicted bounding box and the annotation bounding box intersect, as the vehicles are extremely small.


 

\tabcolsep=0.5pt
	   	 
             


\begin{table*}[t]
  \centering
  \renewcommand\tabcolsep{5pt}
   \caption{Quantitative results of different datasets and methods.}
   \vspace{-3mm}
   \resizebox{0.85\linewidth}{!}{
  \begin{tabular}{c|ccc|ccc|ccc}
    \hline
     \multirow{3}{*}{Method}& \multicolumn{6}{c|}{Training on \textbf{VISO}} & \multicolumn{3}{c}{Training on \textbf{SDM-Car}}\\\cline{2-10}
      ~& \multicolumn{3}{c|}{Test on \textbf{VISO}} & \multicolumn{3}{c|}{Test on \textbf{SDM-Car}} & \multicolumn{3}{c}{Test on \textbf{SDM-Car}}\\\cline{2-10} 
     ~ & Recall & Precision & $F_1$ Score & Recall & Precision & $F_1$ Score & Recall & Precision & $F_1$ Score \\
    \hline
    VIBE~\cite{barnich2009vibe}&68.8 & 34.7& 45.7 & \textbf{65.7\textcolor{blue}{~(-3.1)}}& 29.3\textcolor{blue}{~(-5.4)}& 40.5\textcolor{blue}{~(-5.2)}&  65.7\textcolor{red}{~(+0.0)} & 29.3\textcolor{red}{~(+0.0)} & 40.5\textcolor{red}{~(+0.0)}  \\
    GMM~\cite{friedman2013image}& 53.6& 55.2 &54.4 & 46.2\textcolor{blue}{~(-7.4)} & 45.5\textcolor{blue}{~(-9.7)} & 45.9\textcolor{blue}{~(-8.5)} & 46.2\textcolor{red}{~(+0.0)} & 45.5\textcolor{red}{~(+0.0)} & 45.9\textcolor{red}{~(+0.0)}  \\
    KNN~\cite{elgammal2000non} & 62.1 & 58.7 & 60.4 & 58.3\textcolor{blue}{~(-3.8)}  & 41.2\textcolor{blue}{~(-17.5)}  & 48.3\textcolor{blue}{~(-12.1)}  & 58.3\textcolor{red}{~(+0.0)} & 41.2\textcolor{red}{~(+0.0)} & 48.3\textcolor{red}{~(+0.0)} \\
    HB-YOLO~\cite{yu2023hb} & 69.2 & 72.1 & 70.6& 49.1\textcolor{blue}{~(-20.1)} & 55.8\textcolor{blue}{~(-16.3)} & 52.2\textcolor{blue}{~(-18.4)} & 66.8\textcolor{red}{~(+17.7)} & 85.7\textcolor{red}{~(+29.9)} & 76.6\textcolor{red}{~(+24.4)}\\
    DSFNet~\cite{9594855} & 75.4 & \textbf{88.2} & 81.3 & 54.3\textcolor{blue}{~(-21.1)}& 63.5\textcolor{blue}{~(-24.7)}& 58.5\textcolor{blue}{~(-22.8)} &  70.7\textcolor{red}{~(+16.4)} & 88.3\textcolor{red}{~(+24.8)} & 78.5\textcolor{red}{~(+20.0)}\\ \hline
    \textbf{Ours} & \textbf{77.3} & 86.8 &\textbf{81.8} & 58.5\textcolor{blue}{~(-18.8)} & \textbf{63.9\textcolor{blue}{~(-22.9)}} & \textbf{61.1\textcolor{blue}{~(-20.7)}} & \textbf{76.8\textcolor{red}{~(+18.3)}} & \textbf{89.6\textcolor{red}{~(+25.7)}} & \textbf{82.7\textcolor{red}{~(+21.6)}} \\ \hline
  \end{tabular}}
 \vspace{-4mm}
  \label{tab:baselines}
\end{table*}



\subsection{Quantitative Results}
\subsubsection{Evaluation of proposed dataset} We first train all baseline methods on the VISO dataset, and test them on the VISO and SDM-Car test sets, respectively. The results shown in Table~\ref{tab:baselines} indicate that all methods suffer performance degradation on SDM-Car test set compared to the VISO test set. This implies that the model trained on the VISO dataset lacks the ability to fit dim vehicles, making it challenging to detect a large number of dim vehicles in the SDM-Car dataset. In addition, we trained all baseline methods on the SDM-Car dataset, and tested them on SDM-Car test set to explore the improvement of model performance training on datasets rich in dim instances. The results in Table~\ref{tab:baselines} show that the proposed SDM-Car dataset does boost the capacity of learning-based baseline modes to detect dim vehicles, with the recall, precision and $F_{1}$ score increased by 17.5, 26.8 and 22.0 on average, respectively.

\begin{figure}[tb]
    \centering
    \includegraphics[width=0.99\linewidth]{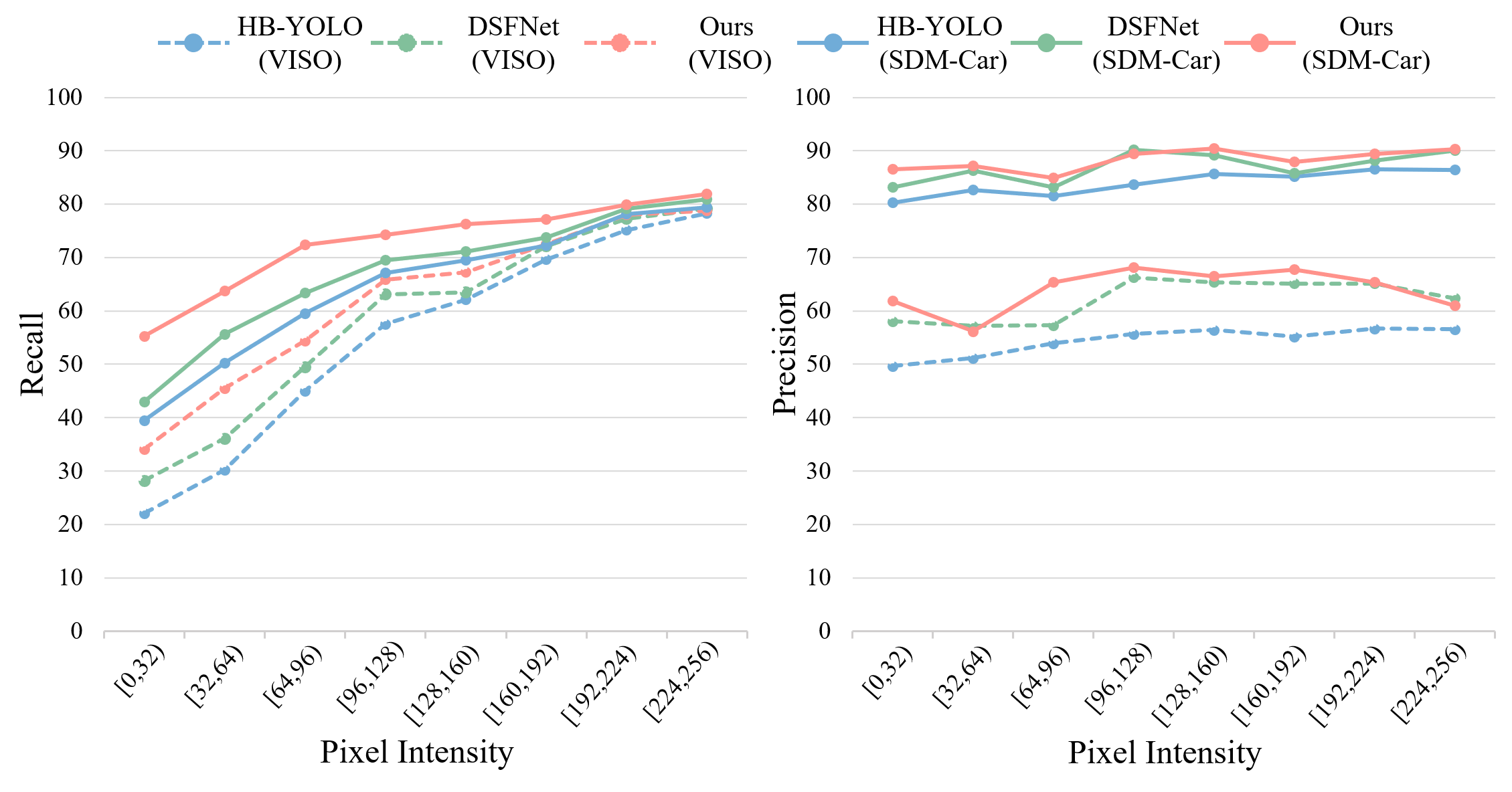}
    \vspace{-8mm}
    \caption{Fine-grained analysis of performance improvements in different pixel intensity intervals.}
    \vspace{-4mm}
    \label{fig:sub}
\end{figure}

\subsubsection{Evaluation of proposed method} To verify the effectiveness of the proposed method, we also compare the proposed method with other baselines on test set of SDM-Car. The results are shown in Table \ref{tab:baselines}. 
The results of training on both VISO and SDM-Car show a significant improvement from the state-of-the-art methods. In particular, compared with DSFNet, the recall rate is increased by 4.2 and 6.1 training on VISO and SDM-Car, respectively, indicating that the proposed method has a positive effect on the detection of dim vehicles.


\begin{table}[t]
  \centering
   \caption{Ablation studies on image enhancement and attention.}
   \vspace{-2mm}
      \renewcommand\tabcolsep{5pt}
   \resizebox{0.42\textwidth}{!}{
  \begin{tabular}{c|c|c|ccc}
    \hline
    Case & Enhancement & Attention & Recall & Precision & $F_1$ Score \\
    \hline
   1& &  & 70.7& 88.3 & 78.5  \\
    2&\cmark & & 75.2&87.5  & 80.9  \\
    3& & \cmark &75.8 & \textbf{90.2} &82.4\\
   4& \cmark & \cmark & \textbf{76.8}&89.6& \textbf{82.7} \\
    \hline
  \end{tabular}}
 
  \label{tab:f1-scores}
\end{table}

\subsection{Ablation Study}
\subsubsection{Improvement on different pixel intensities} To prove the improvement of the proposed dataset and method for dim targets, we verified the detection performance on vehicles in different pixel intensity ranges. Specifically, we divide annotations into intervals with a step of 32 according to pixel intensity and calculate the recall with the whole predicted labels, respectively. Similarly, we divide the predicted labels into intervals according to pixel intensity and calculate the accuracy with the whole annotations, respectively. All methods are trained on VISO and SDM-Car and tested on SDM-Car. The results are shown in Fig.~\ref{fig:sub}. In the low pixel intensity intervals, our dataset and method significantly improve the recall rate, while in the high pixel intensity intervals, the difference is not obvious. In addition, the accuracy is significantly improved across the entire interval by training on our dataset. This proves that the proposed dataset and method do alleviate the problem of difficult detection of dim vehicles.

\subsubsection{Analysis of proposed module} 
\begin{figure}[tb]
    \centering
    \includegraphics[width=0.99\linewidth]{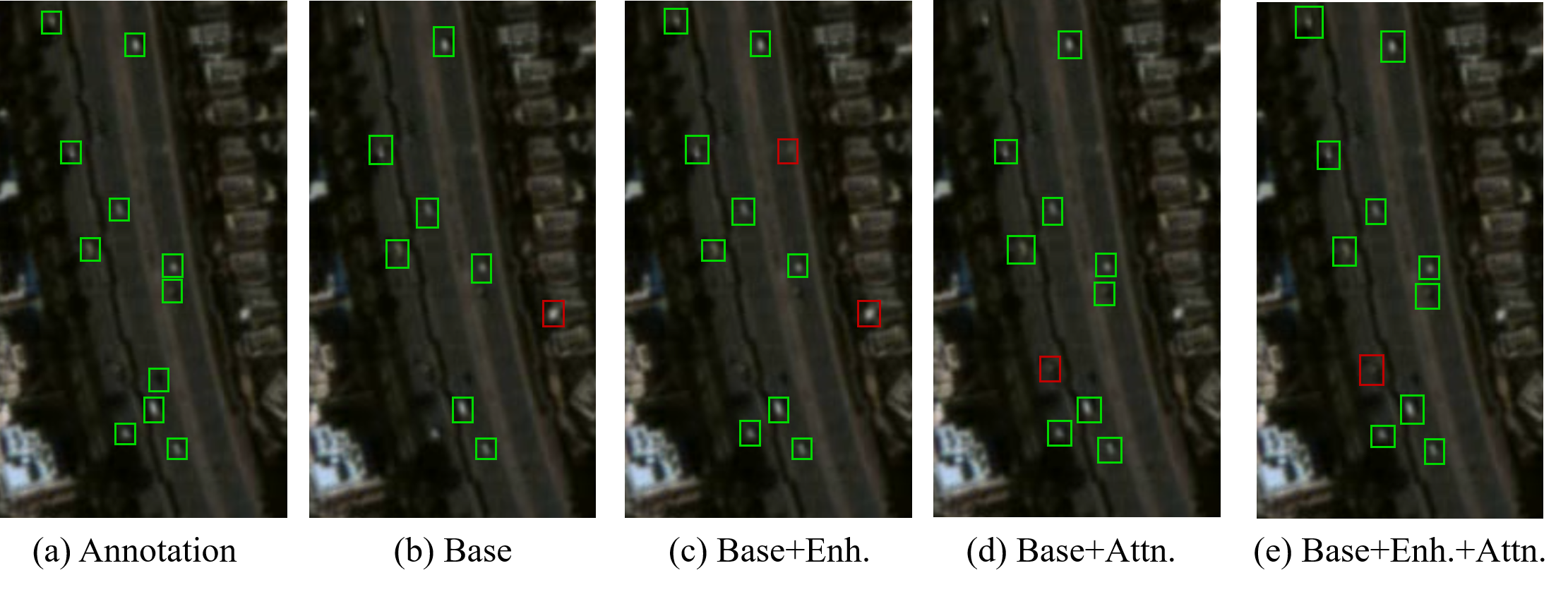}
    \vspace{-8mm}
    \caption{Qualitative analysis on enhancement and attention module. Correct and incorrect detection results are marked with green and red boxes, respectively.}
    \vspace{-5mm}
    \label{fig:qu}
\end{figure}

We investigate the contribution of the modules proposed in our model, that is, the image brightness and contrast enhancement module, and attention module. The results of the quantitative and qualitative experiments are shown in Table III and Fig.~\ref{fig:qu}, respectively. Image enhancement enables model to detect more targets because it can increase the visual appearance of images, making it less difficult to recognize dim targets in complicated backgrounds, although the accuracy is reduced. Attention module can improve both the recall rate and the accuracy, and is able to identify slight vehicle motion on the road background, but it considers objects on complex backgrounds with low confidence. The combination of two strategies leads to simultaneous improvements in both recall and precision and achieves the highest F1 score.

\section{CONCLUSION}
In this paper, we introduced a satellite video dataset named SDM-Car dataset for small and dim vehicle detection and tracking. The SDM-Car dataset consists of 99 annotated videos captured by the Luojia 3-01 satellite. This dataset presents a more challenging task, which aims to facilitate the advancement of the field of small object detection. In addition, to improve the detection of dim targets, we utilized a strategy involving the enhancement of image contrast and brightness and put forward an improved DSFNet that incorporates a SimAM structure. The experimental results show that the proposed dataset and method can significantly improve the detection ability of the model for small and dim vehicles. 
\label{sec:majhead}

\bibliographystyle{IEEEtran}
\bibliography{ref}

\end{document}